\documentclass[10pt,twocolumn,letterpaper]{article}

\usepackage{cvpr}
\usepackage{times}
\usepackage{epsfig}
\usepackage{graphicx}
\usepackage{amsmath}
\usepackage{amssymb}
\usepackage{enumitem}

\DeclareMathOperator*{\argmin}{arg\,min}
\newcommand{\proj}{\operatorname{proj}}
\newcommand{\corr}{\operatorname{ds}}
\newcommand{\DM}{\mathcal{D}}

\newcommand{\suchthat}{\text{s.t.}}
\newcommand{\Refs}{\mathcal{R}}
\newcommand{\Init}{\mathcal{I}}
\newcommand{\GT}{\text{GT}}

\usepackage[usenames,dvipsnames]{color}
\usepackage{xcolor}
\usepackage[font=small]{caption}
\usepackage{booktabs}
\usepackage[margin=5pt]{subfig}
\usepackage{adjustbox}
\usepackage{array}

\newcolumntype{R}[2]{%
    >{\adjustbox{angle=#1,lap=\width-(#2)}\bgroup}%
    l%
    <{\egroup}%
}

\usepackage[pagebackref=false,breaklinks=true,letterpaper=true,colorlinks,bookmarks=false]{hyperref}

\cvprfinalcopy 

\def\cvprPaperID{1437} 

\ifcvprfinal\fi
\begin{document}

\title{Efficiently Creating 3D Training Data for Fine Hand Pose Estimation}

\author{Markus Oberweger \qquad Gernot Riegler \qquad Paul Wohlhart \qquad Vincent Lepetit\\
Institute for Computer Graphics and Vision\\
Graz University of Technology, Austria\\
{\tt\small \{oberweger,riegler,wohlhart,lepetit\}@icg.tugraz.at}
}

\maketitle


\begin{abstract}
  While many recent  hand pose estimation methods critically rely  on a training
  set of  labelled frames, the  creation of such a dataset is a challenging task
  that has been  overlooked so far.  As a result,  existing datasets are limited
  to a few sequences and individuals,  with  limited accuracy, and this prevents
  these methods from delivering their full potential. We propose a semi-automated
  method for  efficiently and accurately  labeling each  frame of a  hand depth
  video with the corresponding 3D locations of  the joints: The user is asked to
  provide only an estimate of the  \emph{2D reprojections} of the visible joints
  in some  reference frames,  which are automatically  selected to  minimize the
  labeling work by  efficiently optimizing a sub-modular  loss function.  We
  then exploit spatial, temporal, and appearance constraints to retrieve the full 3D poses of
  the hand over  the complete sequence.  We show that this  data can be
  used to train a recent state-of-the-art hand pose estimation method, leading to 
  increased accuracy. The code and dataset can be found at \small\url{https://github.com/moberweger/semi-auto-anno/}.
\end{abstract}


\section{Introduction}

Recent             work              on             articulated             pose
estimation~\cite{Ionescu2014,Oberweger2015a,Sun2015,Tang2014,Tompson2014}  has   shown  that  a
large  amount of  accurate training  data  makes reliable and precise estimation 
possible.  For human bodies, Motion  Capture~\cite{Ionescu2014} can be used to
generate large  datasets with  sufficient accuracy.  However, creating accurate
annotations  for hand  pose  estimation  is far  more  difficult,  and still  an
unsolved problem.  Motion Capture  is not  an option  anymore,  as it  is not 
possible to use fiducials  to track the joints of a hand.  Moreover, the  human hand  has more
degrees of freedom than are generally considered for 3D  body tracking, and
an even larger amount of training data is probably required.

\begin{figure}
 \subfloat[]{\includegraphics[width=0.33\linewidth]{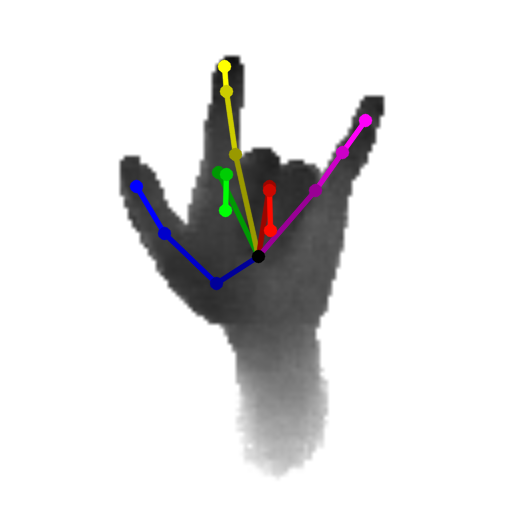}}
 \subfloat[]{\includegraphics[width=0.33\linewidth]{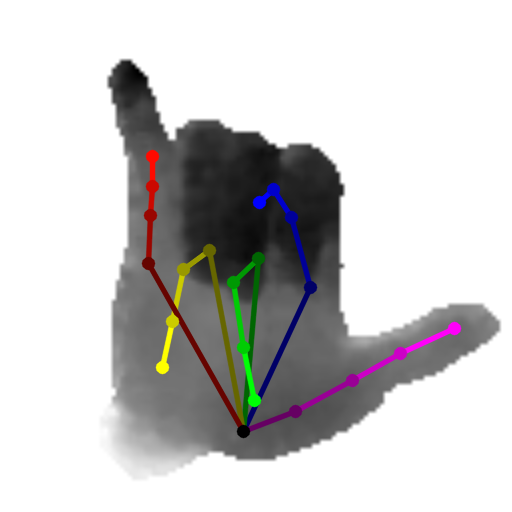}}
 \subfloat[]{\includegraphics[width=0.33\linewidth]{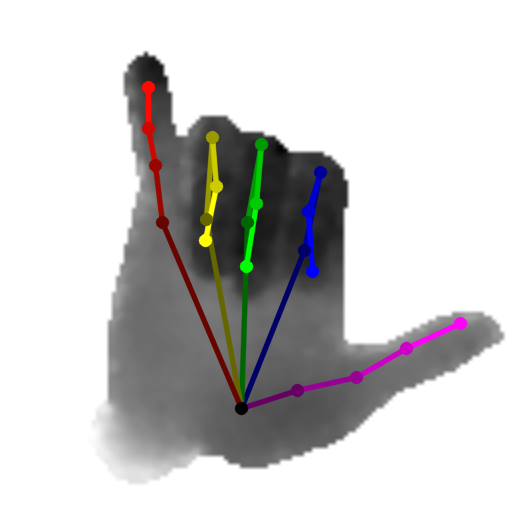}}
 \caption{Recent  hand  pose  datasets  exhibit significant  errors  in  the  3D
   locations of the  joints.  (a) is from the  ICVL dataset~\cite{Tang2014}, and
   (b) from  the MSRA dataset~\cite{Sun2015}.   Both datasets were  annotated by
   fitting a 3D  hand model, which is  prone to converge to a  local minimum. In
   contrast, (c) shows the annotations acquired with our proposed method for the
   same frame as in (b). (Best viewed in color)}
 \label{fig:errors}
\end{figure}

The appearance of depth sensors has made 3D hand pose estimation easier, but has 
not solved the problem of the creation  of training data entirely. Despite its importance,
the creation of  a training set has  been overlooked so far, and  authors have had 
to    rely      on      \emph{ad       hoc}      ways that are prone to
errors, as shown in Fig.~\ref{fig:errors}.       Complex      multi-camera
setups~\cite{Ballan2012,Sridhar2013,Tompson2014,Tzionas2013}    together    with
tracking  algorithms  have  typically  been used  to  create  annotations.   For
example, Tompson~\etal~\cite{Tompson2014} used a complex camera setup with three RGBD
cameras to  fit a  predefined 3D  hand model. Looking  closely at  the resulting
data,  it seems  that  the 3D  model  was  often manually  adjusted  to fit  the
sequences  better and in between these manually adjusted frames the fit can be poor.
Further,  the dataset  of~\cite{Tang2014}  contains  many
misplaced   annotations,   as  discussed   by~\cite{Oberweger2015,Supancic2015}.
Although recent datasets~\cite{Sun2015} have paid more attention to high quality
annotations, they still contain annotation errors, such as multiple annotations on a
single  finger, or  mixing  fingers. These errors  result in  noisy  training and  test data,  and  make  training  and evaluating  uncertain.  This issue  was addressed recently by~\cite{Belagiannis2015}, which shows that  using a robust loss function for training rather 
than a least-squares one results in better performance.

These   problems    can   be   circumvented   via   using   synthetic   training
data~\cite{Riegler2015,Rogez2015,Xu2013}.  Unfortunately, this  does not capture
the sensor  characteristics, such as  noise  and  missing data typical  of depth
sensors, nor  the physical  constraints that  limit the  range of  possible hand
poses~\cite{Wu2001}. Another common approach to creating training data is using
crowd source platforms,  such as Amazon Mechanical Turk.  In  our case, however,
the annotations should be  in 3D, which makes the task  very challenging if done
manually, even with a depth sensor: The sensor can only provide the depth of the
skin,  not the  joints  themselves,  and even  this  information  is not  always
available in the case of self-occlusion or missing data. Thus, this task does not 
lend itself to this kind of crowd  sourcing  with untrained  workers.  Moreover,
whatever the method, one has to recreate new data for each new sensor.

For all of these reasons, we  developed a semi-automated  approach that makes it
easy to  annotate  sequences  of  articulated  poses  in  3D.  We  ask  a  human
annotator to provide  an estimate of the 2D reprojections  of the visible joints
in  frames  we  refer to as \emph{reference frames}.  We  propose  a  method  to
automatically select these  reference frames to minimize  the annotation effort,
based on  the appearances of  the frames over the  whole sequence.  We  then use
this information to  automatically infer the 3D locations of  the joints for all
the frames, by  exploiting appearance, temporal, and  distances constraints.  If
this inference fails for some frames, the annotator can still provide additional
2D reprojections;  by running  the global inference  again, a  single additional
annotation typically fixes many frames.

\begin{figure}
  \includegraphics[width=0.16\linewidth]{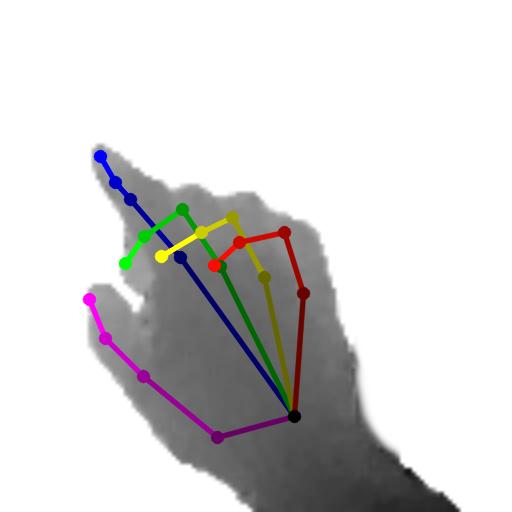} \hspace*{-0.2em}
  \includegraphics[width=0.16\linewidth]{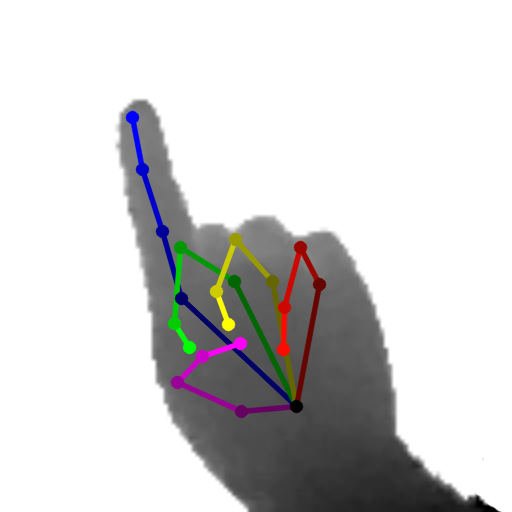} \hspace*{-0.2em}
  \includegraphics[width=0.16\linewidth]{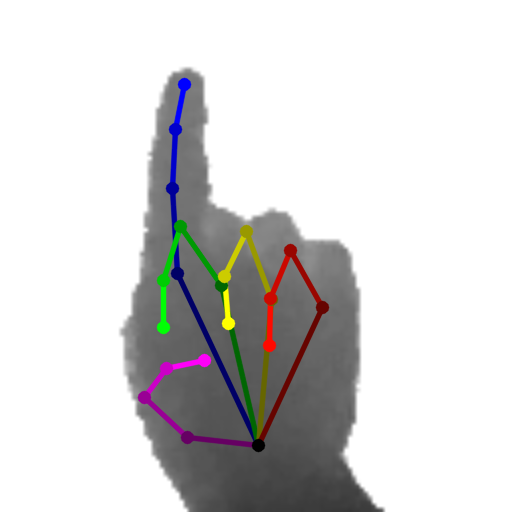} \hspace*{-0.2em}
  \includegraphics[width=0.16\linewidth]{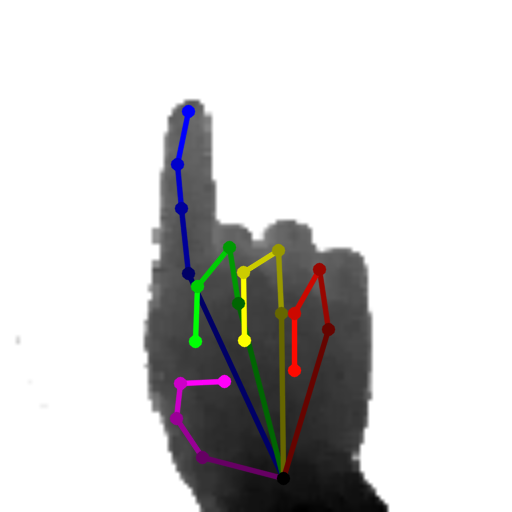} \hspace*{-0.2em}
  \includegraphics[width=0.16\linewidth]{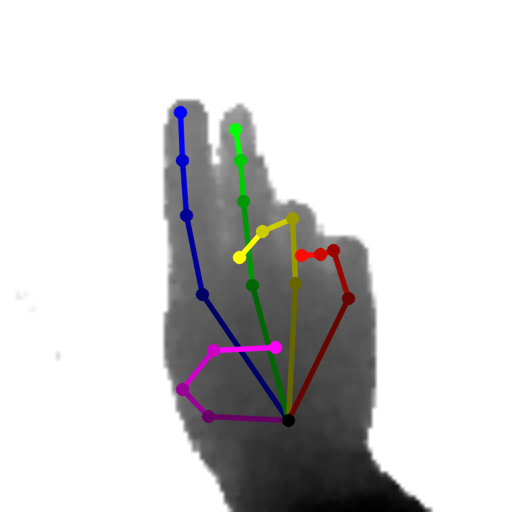} \hspace*{-0.2em}
  \includegraphics[width=0.16\linewidth]{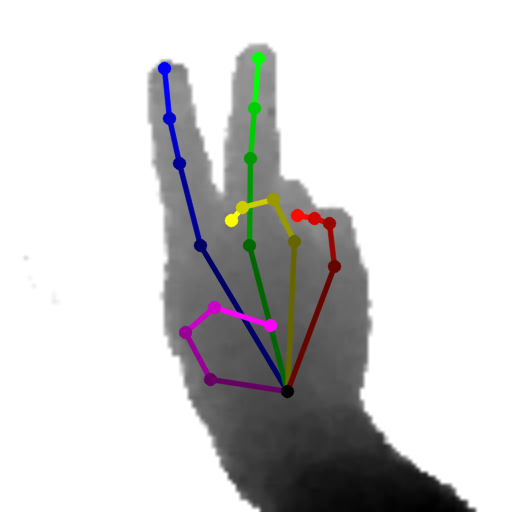} \hspace*{-0.2em}

  \includegraphics[width=0.16\linewidth]{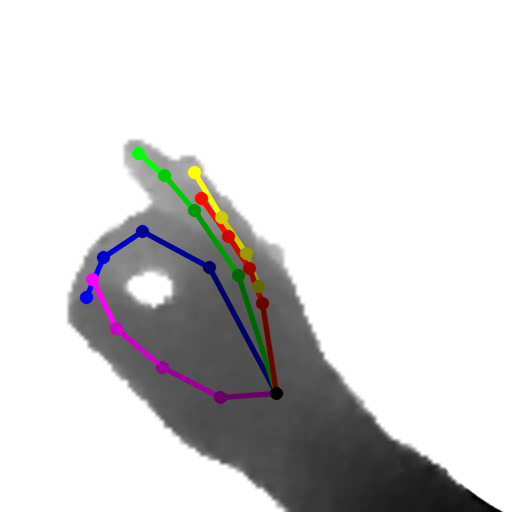} \hspace*{-0.2em}
  \includegraphics[width=0.16\linewidth]{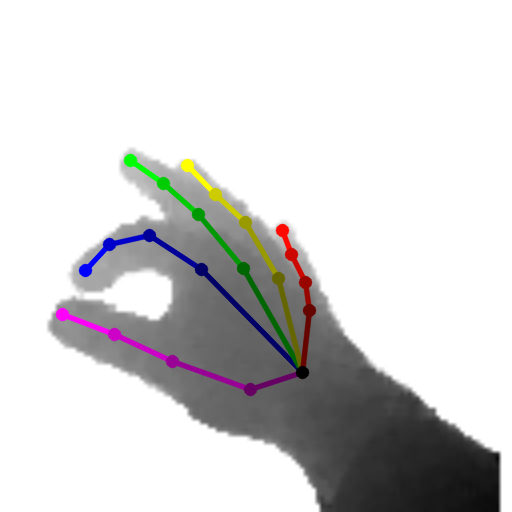} \hspace*{-0.2em}
  \includegraphics[width=0.16\linewidth]{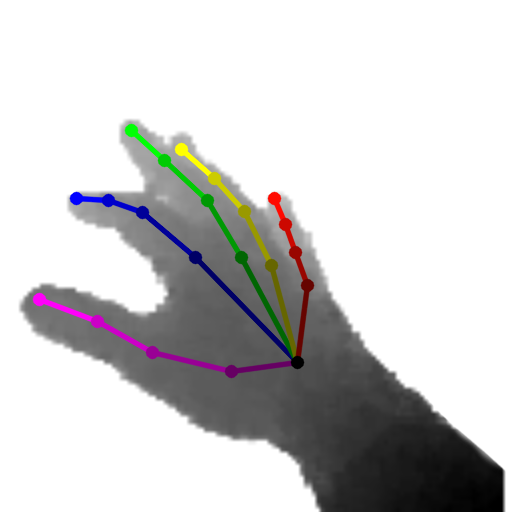} \hspace*{-0.2em}
  \includegraphics[width=0.16\linewidth]{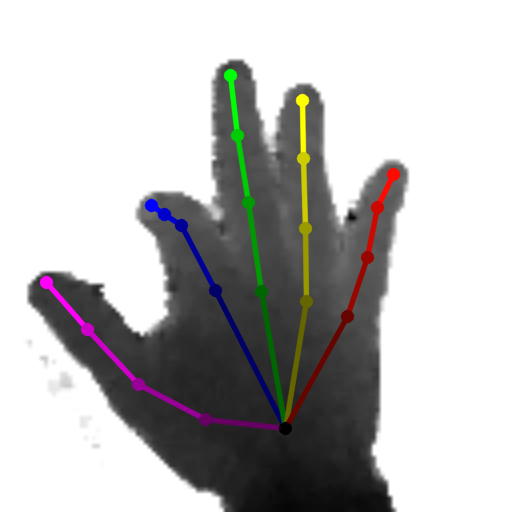} \hspace*{-0.2em}
  \includegraphics[width=0.16\linewidth]{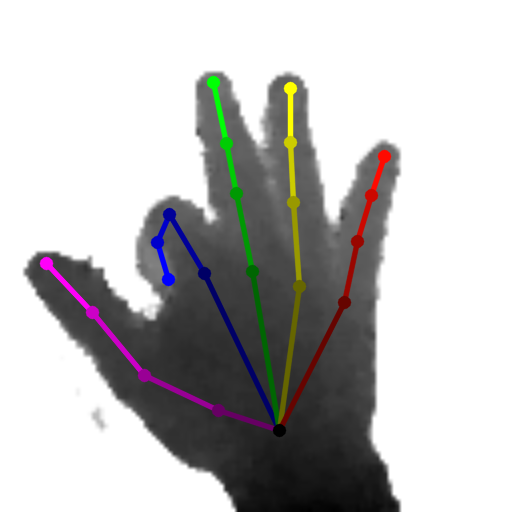} \hspace*{-0.2em}
  \includegraphics[width=0.16\linewidth]{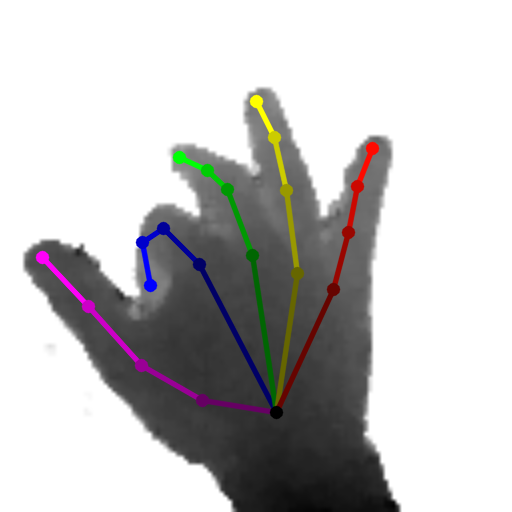} \hspace*{-0.2em}
  \caption{Our method made it possible to create a fully annotated dataset of more than
    2000 frames  from an egocentric  viewpoint, which  is considered to  be very
    challenging~\cite{Rogez2014}.  (Best viewed in color)}
  \label{fig:ego}
\end{figure}

We evaluate our approach using both synthetic data and real images. We also show
that  we can  improve the  annotations of  existing datasets,  which yield  more
accurate predicted poses.  As Fig.~\ref{fig:ego} shows, our approach also allows
us to  provide the first  fully annotated  egocentric sequences, with  more than
2000 frames in total. We will make this sequences and the full code available on our website.


\section{Related Work}
\label{sec:relatedwork}

Complex camera setups can be used to mitigate problems with self-occlusions.
Tompson~\etal~\cite{Tompson2014}  relied on  three  RGBD cameras.   They used  a
predefined 3D  hand model that  had to be  manually readjusted for  each person.
When looking closely  at the data, it  appears that the dimensions  of the model
were  modified  over  the  sequences,   probably  to  fit  the  incoming  images
better. This dataset was taken from a frontal view of the user, which limits the
range of the poses.  Sridhar~\etal~\cite{Sridhar2013} used five RGB and two RGBD
cameras, and annotated only the finger tips, which is not enough for full 3D pose
estimation. \cite{Ballan2012,Tzionas2013} required eight  RGB cameras to capture
hand interactions,  however, causing significant  restrictions on  hand
movement within this setup.

An alternative to these complex setups  with restricted ranges are single camera
approaches.    For   example,   Tang~\etal~\cite{Tang2014}   used   the   method
from~\cite{Melax2013} to fit  a hand model to a single  depth image.  Similarly,
\cite{Qian2014,Sun2015} used a  single depth camera to fit a  predefined 3D hand
model. These methods are based  on frame-to-frame tracking. This requires manual
supervision, and  leads to  many errors  if the  optimization does  not converge
correctly.

Very accurate training data can be generated using synthetic models, as was done
in  \cite{Riegler2015,Rogez2015,Xu2013} for  example.   However, synthetic  data
does not  capture the full  characteristics of the  human hand, and  sensor
characteristics are not considered.  \cite{Xu2013} added synthetic sensor noise,
however, it is difficult to model this in a general way.

There  are also  invasive  methods  for acquiring  accurate  3D locations.   For
example, \cite{Wetzler2015} used  a sophisticated magnetic tracker  but only for
finger tips.  \cite{Xu2013} used a data glove, but unfortunately data gloves are not
very  accurate, and  would  be  visible in  the  training  images, thus  biasing
learning algorithms.

A different approach was proposed by Yasin~\etal~\cite{Yasin2016}, who matched 2D poses against a set of 3D poses obtained from
motion capture  sequences, by comparing the  2D poses with the reprojections of
the 3D  poses in virtual  cameras. This is an interesting approach, however,  2D pose
estimation is also an open research topic and still prone to errors.

For egocentric 3D hand pose annotation, Rogez~\etal~\cite{Rogez2014} proposed a semi-automatic labeling  method, where a user labels
the 2D locations of a few joints, and chooses the closest 3D pose among a set of
synthetic  training  samples.  The  3D  pose  is  then  estimated  from  the  2D
annotations and  the selected 3D training  pose.  The user then  has to manually
refine the pose  in 3D. This process  is iterated until an  appealing result is
achieved.   This is  a time  consuming  task and  thus, they  only created  a
temporally sparse  set, which  is only  sufficient  for testing  and additional  
data  is required  for training.

Semi-automated methods for  annotating video sequences like ours are  not new to
Computer Vision.   \cite{Lepetit2000} exploited object silhouettes  in reference
frames to predict the object  silhouettes in the remaining frames.  \cite{Ali11}
also used  manual annotations of  some frames to  iteratively train a  2D object
detector. \cite{Wei2010} used annotations in manually selected frames, to predict the  annotations of the remaining ones. Compared to these works, we propose a method for selecting the frames to
be  annotated, minimize  manual  work, but  more  importantly, our  approach
provides a complex articulated 3D structure from 2D annotations.


\section{Creating Training Data Efficiently}
\label{sec:main}

Given a sequence of $N$ depth maps $\{\DM_i\}_{i=1}^N$ capturing a  hand in motion,
we want to estimate the 3D joint locations for each $\DM_i$ with minimal effort.
Our approach  starts by automatically selecting  some of the depth  maps we will
refer  to as \emph{reference frames} (Section~\ref{sec:ref}). A user  is then  asked to  provide the  2D
reprojections of the joints in these reference frames, from which we infer their
3D locations  in these  frames (Section~\ref{sec:init}).  We  propagate these 3D  locations to  the other
frames (Section~\ref{sec:propagate}),  and  we  perform  a  global  optimization,  enforcing  appearance,
temporal, and spatial constraints (Section~\ref{sec:global}).

\subsection{Selecting the Reference Frames}
\label{sec:ref}

A simple  way to select  the reference frames would  be to regularly  sample the
video sequence  in time, and select, for example, every tenth frame as reference 
frame. However,  this solution  would be  sub-optimal: Sometimes  the fingers
move fast,  and a higher  sampling rate would be  required, while they  can also
move  more  slowly,  requiring less  manual  annotation.   Moreover, hand motion
performers tend to move back to similar poses at wide intervals, and annotating
the same poses several times should be avoided.

Simple temporal sampling therefore does not seem to be a good approach. Ideally, we
would like to select as few reference frames as possible, while making sure that
for each frame,  there is at least  one reference frame that  is similar enough.
This  will  ensure that  we  can  match them  together  and  estimate the  joint
reprojections in  the frame.   Let us  assume that we  know a  distance function
$d(\DM_i, \DM_j)$ that can be used  to evaluate the similarity between two depth
maps $\DM_i$ and $\DM_j$. Then, the  reference frame selection can be formulated
as the following Integer Linear Problem~(ILP):
\begin{eqnarray}
  \argmin_{\{x_i\}_{i=1}^N} \sum_i^N x_i \quad \suchthat \quad \forall i \> \sum_{j \in E_i} x_j \geq 1 \> , \\
\text{with }  E_i = \{j \; | \; d(\DM_i,\DM_j) \leq \rho\} \> ,
\end{eqnarray}
where the $x_i$ indicate which frames are selected as reference frames ($x_i =
1$ if $\DM_i$ is selected, and 0 otherwise). $E_i$ is the set of indices of the
frames that are similar enough to  frame $i$ for matching, according to distance
function $d(\cdot,\cdot)$, and $\rho$ is a threshold.

This formulation  guarantees that we find the global optimum. We  implemented it
using~\cite{Berkelaar2005}  but  unfortunately, optimization  turned  out to  be
intractable for real problems with the number of frames $N$ larger
than about $10^3$. Thus, we turned to the suboptimal but tractable approach by optimizing:
\begin{equation}
  \max_\Refs \; f(\Refs) \quad \suchthat \quad |\Refs| < M \> ,
\end{equation}
where $\Refs$ is the  set of selected reference frames, $M$ the maximum number of reference frames, and $f(\Refs)$  is the number of
frames within the chosen  distance $\rho$ to at least one of the  frames in $\Refs$. In
this approach,  if $M$ is set  too small, some  frames may not have  a reference
frame near them, but we can trade  off the coverage by reference frames with the
amount of annotation work. This optimization problem is a submodular problem and
it  is  NP-complete, but  what  makes  it attractive  is  that  a simple  greedy
optimization   was   shown   to  deliver a solution that is close to the  globally 
optimal one~\cite{Nemhauser1978}. This greedy optimization procedure simply
proceeds by adding the element $e$ to the set $\Refs$ that maximizes the difference 
$f(e \cup \Refs) - f(\Refs)$ as long as the number of reference frames is smaller than $M$.

\begin{figure}
  \begin{center}
    \includegraphics[width=\linewidth]{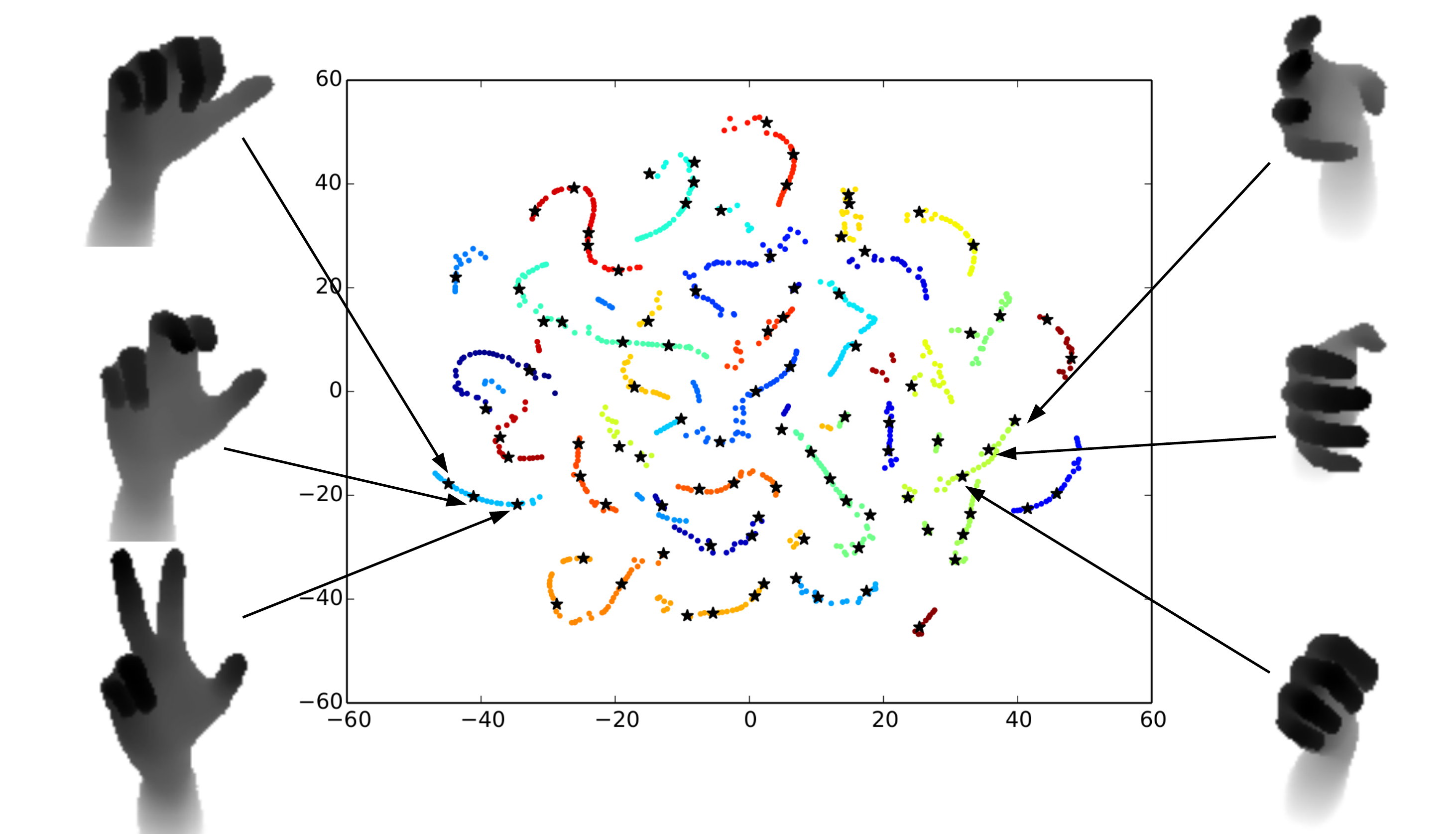}
  \end{center}
  \caption{t-SNE~\cite{Maaten2008}  visualization  of  the  depth  map
    embedding over a sequence. Each colored dot $\bullet$ represents a
    frame, the color encodes the  temporal order.  Temporal changes of
    the hand articulations can be  clearly observed from the different
    trajectories.   The  reference frames  are  shown  as black  stars
    $\bigstar$.   We   automatically  select   them   so   that  their
    annotations can be propagated to the other frames while minimizing
    the manual annotation effort.  The selected reference frames cover
    a maximum of other frames within a distance $\rho$, based on their
    appearance. Note that t-SNE sometimes  moves points far apart that
    are close  to each other  in the original  space. This is  why the
    points do not  form a continuous curve even if  they correspond to
    consecutive frames~\cite{Maaten2008}. (Best viewed on screen)}
  \label{fig:reference_frames}
\end{figure}

We define  the distance  function $d(\cdot,\cdot)$  on descriptors  computed for
  depth  maps.    We  tried  LINE-MOD~\cite{Hinterstoisser2011} and
HOG~\cite{Dalal2005}.   However,  the  best  results  were  achieved  by  cosine
distance  between  low  dimensional   embeddings  computed  by  a  convolutional
autoencoder\footnote{Please see  the supplemental  material for the  network 
 architecture.}~\cite{Masci2001}.   Fig.~\ref{fig:reference_frames}  shows
several examples of reference frames selected with this method, visualized along
with the depth map embedding.

\subsection{Initializing the 3D Joint Locations in the Reference Frames}
\label{sec:init}

Once the procedure described in the  previous section has selected the reference
frames, a human annotator  has to label them. The annotator  is only required to
provide the 2D reprojections of the joints with visibility information in each reference frame, and whether
these joints are closer or farther from  the camera than the parent joint in the
hand  skeleton tree.   This can  be done  easily and  quickly, and  we use  this
information  to automatically  recover  the  3D locations  of  the joints. It is
useful to know the  positions of consecutive joints in relation to the camera in
order  to   avoid   possible   mirroring  ambiguities  typical   of  articulated
structures~\cite{Pons-Moll2014,Sminchisescu2003,Taylor2000}.  We  refer to  this 
information as the \emph{z-order} constraint.

To automatically  recover the  3D locations  of the  joints, we optimize the 
following constrained non-linear least squares problem for each reference frame:
\begin{align}
\label{eq:ref_loss} 
\argmin_{\{L_{r,k}\}_{k=1}^K} & \sum_{k=1}^K v_{r,k} \lVert \proj(L_{r,k}) - l_{r,k}\rVert_2^2 \\
  \suchthat \quad & \forall k \:\: \lVert L_{r,k}-L_{r,p(k)}\rVert_2^2 = d_{k,p(k)}^2 \nonumber \\
  & \forall k \:\: v_{r,k}=1 \Rightarrow \DM_r[l_{r,k}] <  z(L_{r,k}) < \DM_r[l_{r,k}]+\epsilon \nonumber \\
  & \forall k \:\: v_{r,k}=1 \Rightarrow (L_{r,k} - L_{r,p(k)})^\top \cdot c_{r,k} > 0 \nonumber \\
  & \forall k \:\: v_{r,k}=0 \Rightarrow z(L_{r,k}) > \DM_r[l_{r,k}] \nonumber
\end{align}
where $r$ is the index of the reference frame. $v_{r,k} = 1$ if the $k$-th joint
is visible in the $r$-th frame, and 0 otherwise. $L_{r,k}$ is the 3D location of
the  $k$-th joint  for the  $r$-th frame.  $l_{r,k}$ is  its 2D  reprojection as
provided by the human annotator. $\proj(L)$  returns the 2D reprojection of a 3D
location. $p(r)$ returns the index of  the parent joint of
the  $k$-th joint  in  the hand  skeleton. $d_{k,p(k)}$  is  the known  distance
between the  $k$-th joint and its  parent $p(k)$. $\DM_r[l_{r,k}]$ is  the depth
value in $\DM_r$ at location $l_{r,k}$. $z(L)$  is the depth of 3D location $L$.
$\epsilon$  is a  threshold used  to define  the depth  interval of  the visible
joints. In practice,  we use $\epsilon = 15$\,mm given  the physical properties of
the hand.  $c_{r,k}$  is equal to the vector $[0,0,-1]^\top$  if the $k$-th joint
is  closer to  the camera  than  its parent  in frame  $r$, and  $[0,0,1]^\top$
otherwise.  $(L_{r,k}  - L_{r,p(k)})$ is  the vector  between joint $k$  and its
parent in this frame.

Together, the terms of Eq.~\eqref{eq:ref_loss} assure that: (1) the bone lengths
of  the skeleton  are respected;  (2) visible  joints are  in range  of observed
depth values; (3) hidden joints are not in front of observed depth values; and (4)
depth order  constraints of  parent joints are  fulfilled.  We  currently assume
that the  lengths $d_{k,p(k)}$ are known.   In practice, we measure  them from a
depth  map of  the hand  with  open fingers  and  parallel to  the image  plane.
It may also be possible to optimize these distances as  they are 
constant over the sequences from the same person.

We optimize this problem  with SLSQP~\cite{Kraft1988}.  Equality constraints are
hard to optimize, so we relax them and  replace the constraints by a term in the
loss function that penalizes constraint  violations.  We use a simple scheduling
procedure to  progressively increase the  weight of this  term. This gives  us a
reasonable initial estimate of the 3D pose of the hand for each reference frame.

\begin{figure}
  \subfloat[2D annotation]{\includegraphics[width=0.33\linewidth]{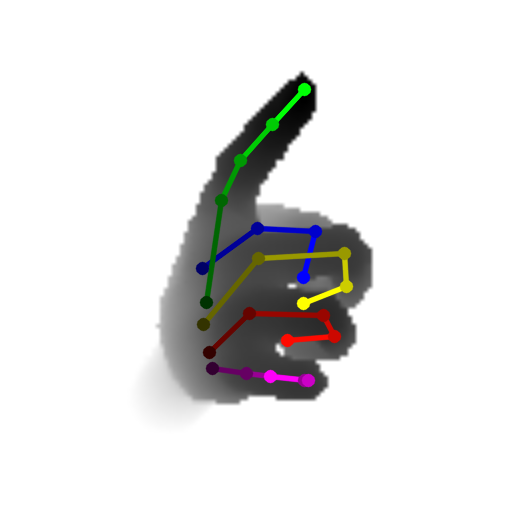}}
  \subfloat[3D initialization]{\includegraphics[width=0.33\linewidth]{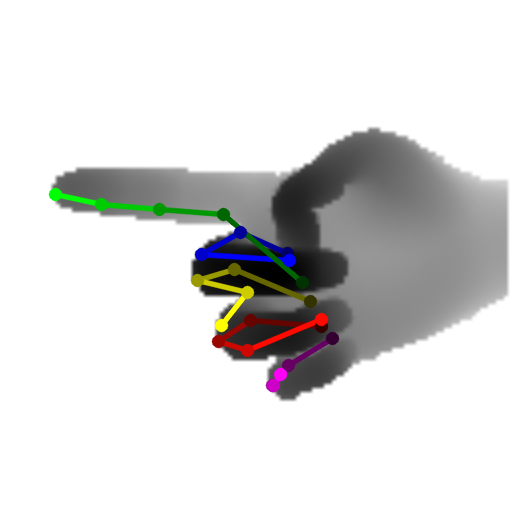}}
  \subfloat[3D result]{\includegraphics[width=0.33\linewidth]{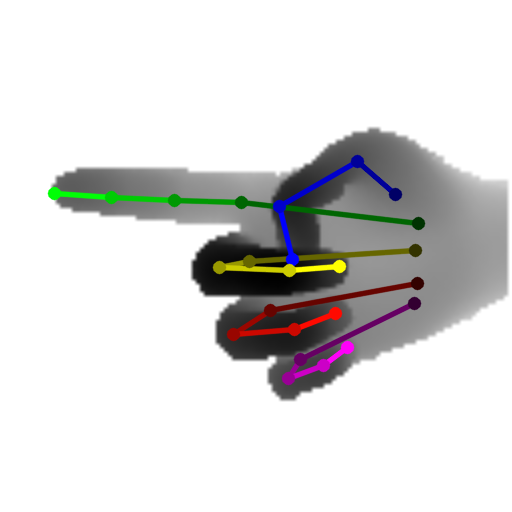}}
  \caption{Optimization  steps  for reference  frames.   We  start with  the  2D
    annotations on a depth image provided by a user (a), and backproject them to
    initialize their  3D location  estimates. (b) shows  the same  pose rendered
    from  a  different viewpoint,  depicting  the  depth initialization  of  the
    joints.  We then  optimize the constrained non-linear least  squares loss of
    Eq.~\eqref{eq:ref_loss} on these initial 3D locations.  The  result is shown
    in  (c), again  rendered from  a different  viewpoint, but  now with  better
    aligned 3D locations. (Best viewed in color)}
  \label{fig:reference_frame_opt}
\end{figure}

We initialize  the joint  depth with  the measurement
from the depth  sensor at the annotated 2D location.  This is shown in
Fig.~\ref{fig:reference_frame_opt}, which  depicts the  initialized 3D
locations and the result after optimizing the relaxation of Eq.~\eqref{eq:ref_loss}.

\subsection{Initializing the 3D Joint Locations in the Remaining Frames}
\label{sec:propagate}

The  previous section  described how  to  compute a  first estimate  for the  3D
locations of  the joints in the  reference frames.  Next, we iteratively propagate these 3D
locations from  the reference frames to  the remaining frames,  in a
way  similar  to~\cite{Kuettel12},  as  explained  in this section.  This gives
us an initialization for the joint locations in all the frames. The next
subsection will explain how we refine them in a global optimization procedure.

$\Init$ is used to denote the  set of frames for  which the 3D locations  of the
joints have already been initialized.   At the beginning, $\Init$ is initialized
to the  set of  reference frames,  but each time  we estimate  the joints  for a
frame, this frame is  added to $\Init$. At each iteration,  a frame $\hat{c}$ not 
yet initialized and its closest frame $\hat{a} \in \Init$ are selected: 

\begin{equation}
  \begin{bmatrix} \hat{c} \\ \hat{a}  \end{bmatrix} = \argmin_{\substack{{c \in [1;N]\backslash\Init}\\{a\in\Init}}}  d(\DM_c, \DM_a) \> .
\end{equation}

We  use   the  appearance of the  joints  in   $\hat{a}$ to predict  their 3D locations $\{L_{\hat{c},k}\}_k$ in
$\hat{c}$ by minimizing:
\begin{align}
  \begin{split}
    \argmin_{\{L_{\hat{c},k}\}_k} &\sum_k \corr(\DM_{\hat{c}}, \proj(L_{\hat{c},k});
    \DM_{\hat{a}}, l_{\hat{a},k})^2 \\
    \suchthat \quad & \forall k \quad \lVert
    L_{\hat{c},k}-L_{\hat{c},p(k)}\rVert_2^2 = d_{k,p(k)}^2 \> ,
  \end{split}
  \label{eq:nonref_loss}
\end{align}
where $\corr(\DM_1, \proj(L_1); \DM_2, l_2)$ denotes the dissimilarity between the patch in
$\DM_1$  centered  on the  projection  $\proj(L_1)$  and  the patch  in  $\DM_2$
centered on $l_2$.  This optimization looks for joints based on their
appearances  in frame  $\hat{a}$ while  enforcing the  3D distances  between the
joints.  We use the  Levenberg-Marquardt algorithm to solve Eq.~\eqref{eq:nonref_loss},
by relaxing  the hard constraint  with a weighted  additional term in  the loss
function.

\begin{figure}
  \subfloat[Closest reference frame]{\includegraphics[width=0.33\linewidth]{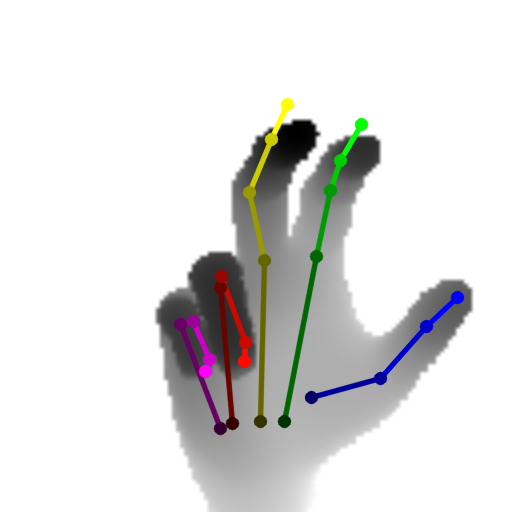}}
  \subfloat[Initialization with SIFTFlow]{\includegraphics[width=0.33\linewidth]{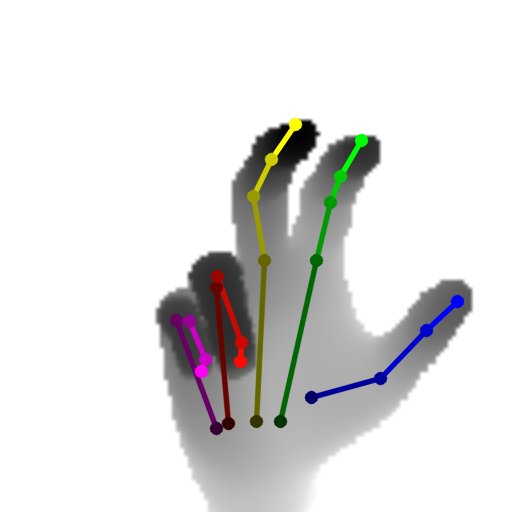}}
  \subfloat[After optimization]{\includegraphics[width=0.33\linewidth]{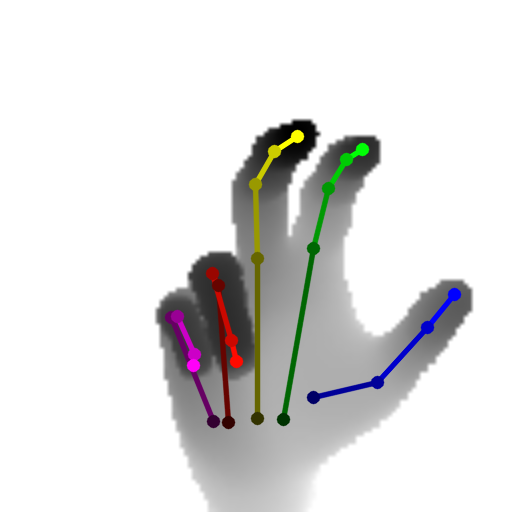}}
  \caption{Initialization of locations  on a non-reference frame.  (a) shows the
    3D  locations for  the  closest reference  frame. (b)  We  propagate the  3D
    locations using SIFTFlow~\cite{Liu2011}. (c) 3D locations after optimization
    of Eq.~\ref{eq:nonref_loss}. (Best viewed in color)}
  \label{fig:init_frames}
\end{figure}

As  illustrated in  Fig.~\ref{fig:init_frames}, we  initialize the  optimization
problem of Eq.~\eqref{eq:nonref_loss} by aligning frames $\hat{c}$ and $\hat{a}$ using
SIFTFlow~\cite{Liu2011}. This maps the 2D reprojections of the joints in frame
$\hat{a}$  to   2D  locations   $\{\tilde{l}_k\}_k$  in  frame   $\hat{c}$.   We
backproject each  $\tilde{l}_k$ on the  depth map $\DM_{\hat{c}}$  to initialize
$L_{\hat{c},k}$.  If the depth information  is missing at $\tilde{l}_k$, we
use the  3D point that  reprojects on $\tilde{l}_k$ and  with the same  depth as
$L_{\hat{a},k}$.

\subsection{Global Optimization}
\label{sec:global}

The previous optimization already gives already a good estimate for the 3D locations of
the joints in all frames. However, each frame is processed independently. We can
improve  the estimates  further by  introducing temporal  constraints on  the 3D
locations.  We therefore perform a global optimization over all the 3D locations
$L_{i,k}$ for all the frames by minimizing:
\begin{align}
  \sum_{i\in[1;N]\backslash\Refs} \sum_k & \! \corr(\DM_i, \proj(L_{i,k}); \DM_{\hat{i}}, l_{\hat{i},k})^2 \; + \tag{C} \nonumber \\
  \lambda_M \sum_i \sum_k & \; \lVert L_{i,k} - L_{i+1,k} \rVert_2^2 \; + \tag{TC}  \nonumber \\ 
  \lambda_P \sum_{r\in \Refs} \sum_k & \; v_{r,k} \lVert \proj(L_{r,k}) - l_{r,k}\rVert_2^2 \; \tag{P} \nonumber \\
  \suchthat \quad & \forall i, k \quad \lVert L_{i,k}-L_{i,p(k)}\rVert_2^2 = d_{k,p(k)}^2
  \> \nonumber.
\end{align}
The first term (C) sums the dissimilarities of the joint appearances with those in the closest reference frame 
$\hat{i}  =   \argmin_{a\in\Refs} d(\DM_{i}, \DM_a)$ over the non-reference frames $i$.
The  second term (TC) is a  simple 0-th 
order motion model  that enforces temporal smoothness of the  3D locations.  The
last term (P) of the sum ensures  consistency with the manual 2D annotations  for the reference
frames.   $\lambda_M$ and  $\lambda_P$ are  weights to  trade off  the different
terms.

This  is a  non-convex  problem, and  we  use the  estimates  from the  previous
subsection to  initialize it.  This prevents the optimization from falling  into bad
local  minimums.  

This problem has $3KN$ unknowns for $K$ joints and $N$ frames.  In
practice, the number  of unknowns varies from $10^5$ to  $10^7$ for the datasets
we consider in the evaluation.  Fortunately, this is a sparse problem, which can
be efficiently optimized with the Levenberg-Marquardt algorithm.


\section{Evaluation}
\label{sec:eval}

To validate our  method, we first evaluate  it on a synthetic  dataset, which is
the only way to  have depth maps with ground truth 3D locations of the
joints. We  then provide  a qualitative  evaluation on real  images, and  on the
recent MSRA dataset~\cite{Sun2015}. Finally, we show  that we can use our method
to create a large dataset of egocentric annotated frames.

\subsection{Evaluation on Synthetic Data}

We used the publicly available framework of \cite{Riegler2015} to generate synthetic depth images
along with the  corresponding ground truth annotation. The  sequence consists of
3040 frames, and  shows  a  single  hand  performing  various  poses  and  arm
movements. We refer to the supplemental material for videos.

\begin{figure}
  \begin{center}
    \includegraphics[width=\linewidth]{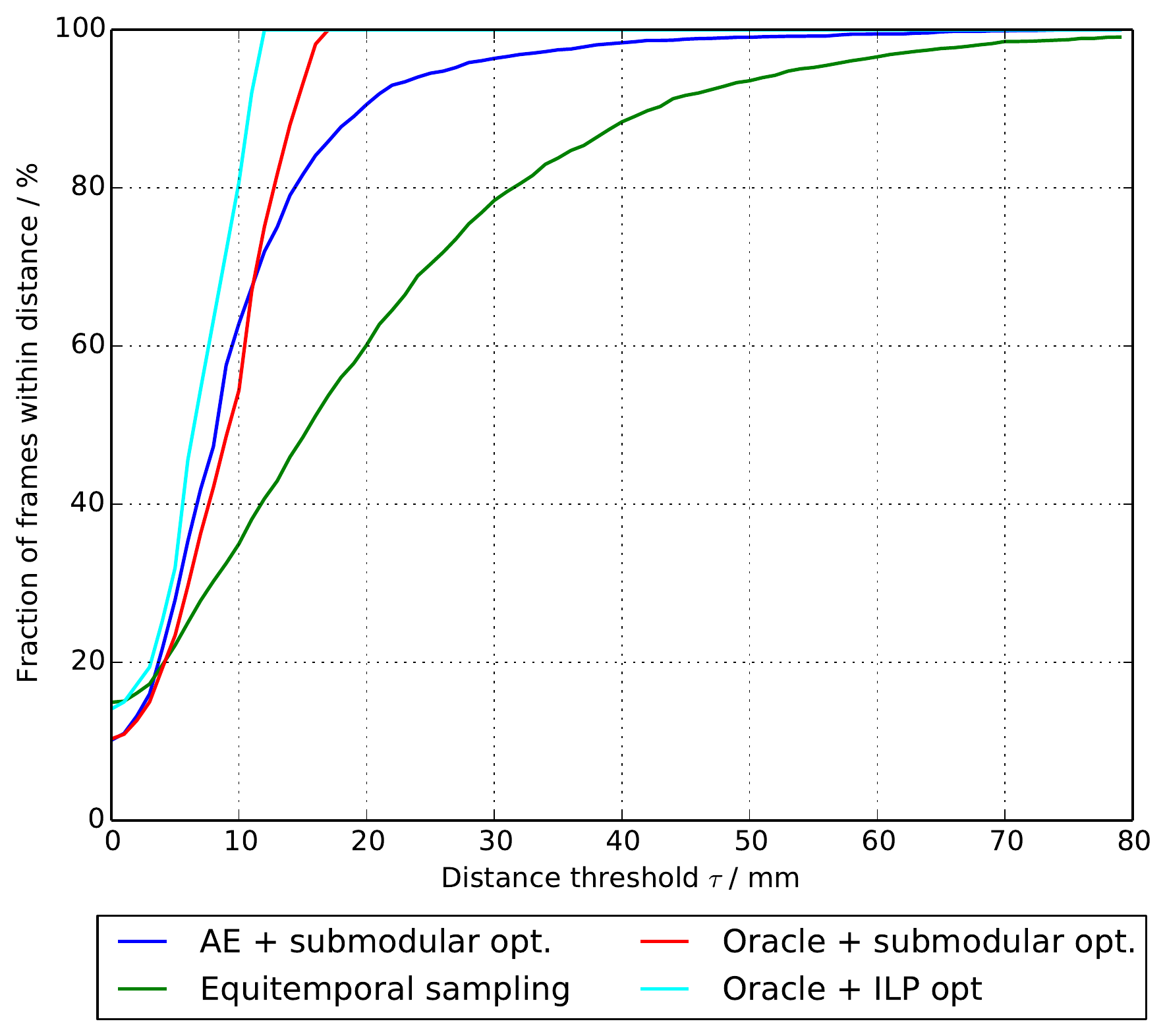}
  \end{center}
  \caption{Evaluation  of the  reference frame  selection method on the
    synthetic sequence.   We
    plot  the  metric  $n(\tau)$ of  Eq.~\eqref{eq:n_metric}  for  our
    selection  of  304  reference  frames,  and  the  same  number  of
    reference frames sampled regularly in  time.  With our method, the
    non-reference frames are closer in 3D to the reference frames. For
    the oracle,  we assume  the pose given,  and apply  the submodular
    optimization  on the  Euclidean pose  distances. The  average pose
    distance of our  proposed selection is only 0.5\,mm  worse than the
    oracle. For  this sequence, the  number of frames is  about $10^3$,
    thus solving  the ILP is  still feasible. The  approximation using
    our submodular optimization is close  to the optimal ILP solution,
    which on average is only 1\,mm better.  (Best viewed in
    color)}
  \label{fig:reference_frameswithin}
\end{figure}

\paragraph{Reference Frame Selection}
In Fig.~\ref{fig:reference_frameswithin},  we plot the fraction  of  frames
for which  the maximum  3D distance of  their joints to  their locations  in the
assigned reference frame is lower than  a threshold.  More formally, we plot the
function $n(\tau)$ with:
\begin{equation}
  n(\tau) := \frac{1}{N}
  \Big| \Big\{ i\in[1;N] \; | \;
  \max_k \;\lVert L^\GT_{i,k} - L^\GT_{a(i),k}\rVert_2 <
  \tau \Big\} \Big| \> ,
  \label{eq:n_metric}
\end{equation}
where the  $L^\GT_{i,k}$ are the ground  truth 3D locations for  the joints, and
$a(i) = \arg\min_{a\in\Refs} d(\DM_i, \DM_a)$.   This metric allows us to check
if the  selection based on  the visual  appearance using distance  $d(\cdot, \cdot)$ also retrieves
reference frames that are close in 3D.  This is an important factor for the
rest of the  method, as the propagation  step will perform better if  a frame is
not too far away from the closest  reference frame.  We use $\rho=0.1$, which we
obtained by  cross-validation,  however, the reference  frame selection  is not
very sensitive to the exact value.

We  compare our  selection with  a  straightforward selection  based on  regular
temporal sampling using the same number  of reference frames.  According to this
metric, our method is significantly better, and it actually yields more accurate
3D  annotations: Using  our proposed  selection, the  average error  is 5.53\,mm,
compared to 6.45\,mm when taking equitemporal samples. Additionally, the required
number of manual reannotations is much higher. Our method required 133 additional 2D
locations, compared to 276 manual interventions for the equitemporal selection.

\paragraph{Reference Frame Initialization}

Table~\ref{tab:ref_init}  provides the  accuracy  of the  initialization of  the
3D joint locations for the reference  frames, and evaluates the influence of the
different  terms  in  Eq.~\eqref{eq:ref_loss}.   We first  use  the  perfect  2D
locations  of  the joints  from  ground  truth.   When  only minimizing  the  2D
reprojection errors with the 3D distance  constraints, the error is quite large.
Adding the  visibility constraint significantly  improves the accuracy,  but the
maximum error is still large.  By adding the z-order term, depth ambiguities due
to mirroring can be resolved, and the errors get much smaller.

\begin{table}
  {\small
  \begin{tabular}{lcc}
    \toprule
    Method & Visible joints & All joints\\ 
     & Avg.~/ median & Avg.~/ median\\
    \midrule
    2D locations & 12.86 / 8.96\,mm & 19.98 / 13.29\,mm\\
    2D \& visibility & 3.94 / 3.18\,mm & 6.20 / 3.41\,mm\\
    2D \& vis \& z-order & \textbf{2.97 / 2.93\,mm} & \textbf{3.65 / 2.98\,mm}\\
      \midrule
      All + 2D noise & \begin{tabular}[x]{@{}c@{}}$3.70\pm 0.71$\,mm /\\ $2.59\pm 0.21$\,mm\end{tabular} & \begin{tabular}[x]{@{}c@{}}$4.29\pm 0.63$\,mm /\\ $2.56\pm 0.23$\,mm\end{tabular}\\
          \bottomrule 
  \end{tabular}
  }
  \caption{Accuracy of reference frame initialization on the synthetic sequence.
    We provide the average and the median Euclidean joint error over all joints.
    The highest accuracy can be achieved by combining all of our proposed clues:
    2D  reprojection errors,  visibility, and  z-order. The  last row  shows the
    robustness of  our method to  noise, after adding  Gaussian noise to  the 2D
    locations.}\label{tab:ref_init}
\end{table}

In practice, the  2D locations of the  joints provided by a  human annotator are
noisy.   Thus,  we    evaluated  the  robustness  of  our  algorithm  by  adding
Gaussian  noise with  zero mean  and  a standard  deviation  of 3\,pixels  to the  2D
locations. For reference, the width of the  fingers in the depth image is around
25\,pixels.  We  show the  average errors in  Table~\ref{tab:ref_init} after  10 random
runs. The  errors are  only 0.7\,mm  larger than without  noise, which  shows the
robustness of our method to noisy annotations. Interestingly, the median error is lower with noisy initialization, which can be attributed to the non-convex optimization. Due to noise, the convergence can lead to different local minima, thus improving some joint estimates, but significantly worsening others.

\paragraph{3D Location Propagation and Global Optimization}
We  evaluate   the  contributions  of   the  different  optimization   steps  in
Table~\ref{tab:acc_stages}.    We   implement   $\corr(\cdot)$   as   normalized
cross-correlation  with a  patch size  of 25\,pixels.   A cross-validation  among
different  correlation methods  and different  patch sizes  has shown  that  our
method  is not  sensitive to  this choice.   Further, we  use $\lambda_M=1$  and
$\lambda_P=100$. The choice of these values is  not crucial for a good result, as
long  as $\lambda_P  > \lambda_M$  to  emphasize  the reprojections  on  the  2D
annotations.

\begin{figure}
  \begin{center}
    \includegraphics[width=\linewidth]{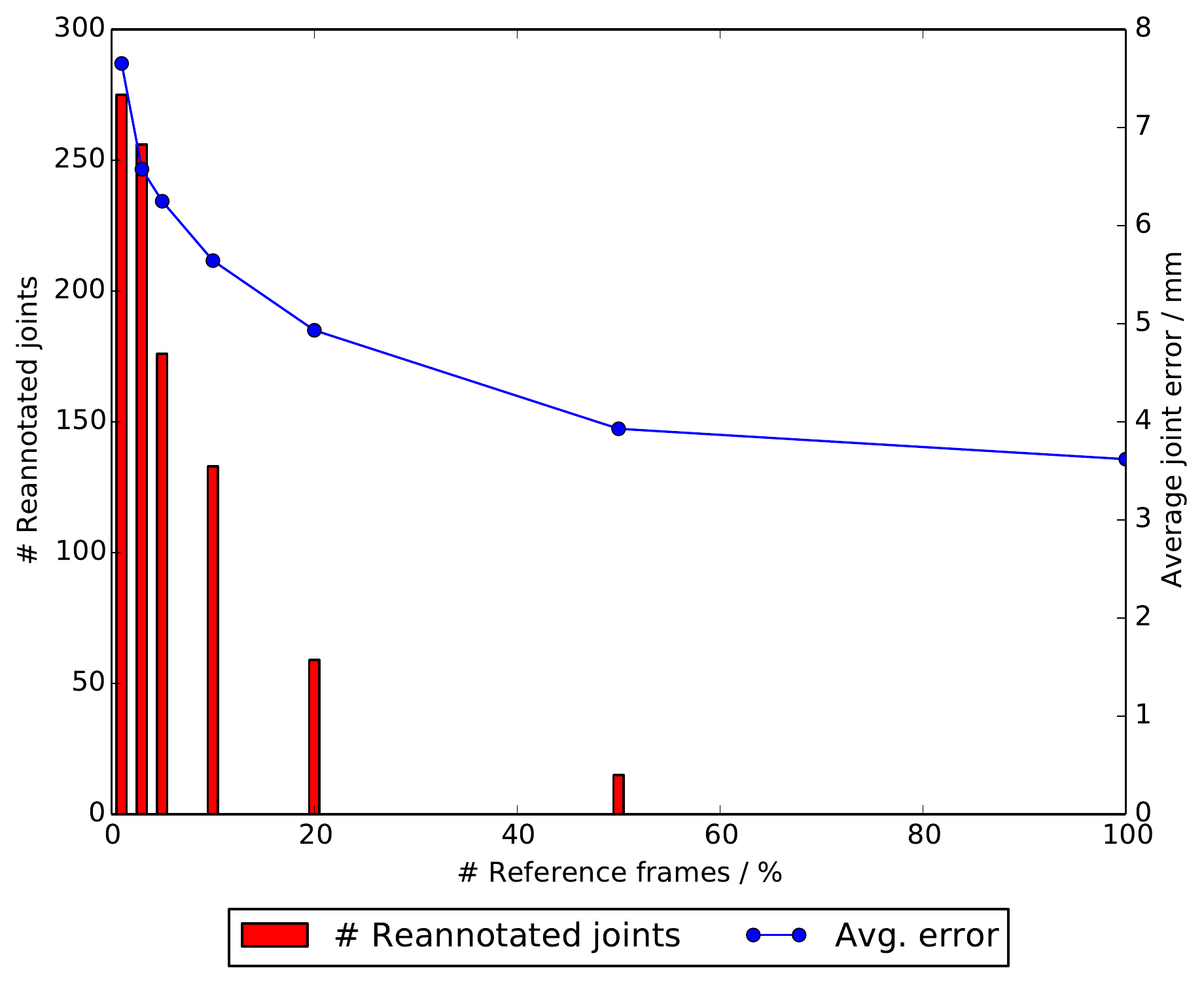}
  \end{center}
  \caption{Accuracy  versus the  number  of reference  frames  on the  synthetic
    sequence.  We  plot the average  3D joint error  and the number  of required
    additional annotations over  the number of the  initially selected reference
    frames. The best  result can be achieved, when  providing manual annotations
    for all frames, however, more reference frames require more annotation work,
    and  for  larger sequences  this  can  be infeasible,~\ie  providing  manual
    annotations for about  $23k$ joints for this sequence.   When decreasing the
    number of initial reference frames, the additional annotations of individual
    joints during  the process increases,  but only in the  hundreds.  Using~\eg
    $3\%$ of all  frames as reference frames requires annotating  only 700 joint
    locations  and  revising  another 250,  while still retaining an  average 3D 
    annotation error of only 6.5\,mm. (Best viewed in color)}
  \label{fig:accuracy_ref}
\end{figure}

If  the  deviation  of the  2D  location  of  a  joint gets  too  large, \ie the 2D location is more than 5\,pixels away from the ground truth location,  manual
intervention is required. For the synthetic  dataset, it was required to readjust
the 2D locations  of 133 joints in 79  frames, or 0.06\% of the  total number of
joints. Fig.~\ref{fig:accuracy_ref}  gives a  more exhaustive evaluation  of the
influence of the chosen  number of reference frames.  We can  obtain a very good
accuracy  by annotating  only a  small percentage  of the  reference frames  and
correcting an even smaller percentage of joints.

\begin{table}
\begin{center}
\small
  \begin{tabular}{lc}
    \toprule
    Method & Avg. / median error\\
    \midrule
    Closest reference & 11.50 / 5.58\,mm\\
    Aligned with SIFTFlow & 11.40 / 5.40\,mm\\
    Frame optimization & 5.76 / 4.34\,mm\\
    \midrule
    Global optimization & 5.53 / 4.23\,mm\\
    \bottomrule 
  \end{tabular} 
  \caption{Accuracy  of the  different  stages on  the  synthetic sequence.   We
    report the  average and  median Euclidean  3D joint errors.   We use  the 3D
    locations of  the reference  frame to initialize  the remaining  frames. The
    first row shows  the accuracy if the 3D locations  of the closest reference
    frame are  used. The next row  shows the contribution of  the alignment with
    SIFTFlow, and  the further optimization on  the 3D locations.  The  last row
    denotes the  accuracy after  the global optimization.  The gain  in accuracy
    with SIFTFlow  is small,  as it  only provides an  offset in  2D, but  it is
    useful to make  the correlation term contribute properly.}\label{tab:acc_stages}
\end{center}
\end{table}

\subsection{Evaluation on Real Data}

\begin{figure}
  \subfloat[Depth camera]{\includegraphics[width=0.49\linewidth]{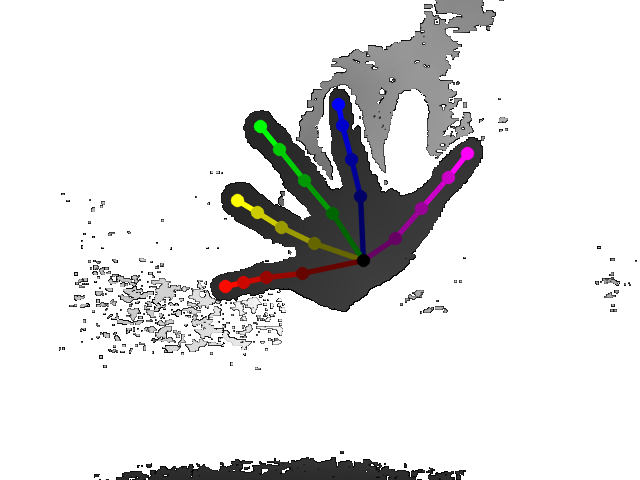}}
  \subfloat[Projected into RGB camera]{\includegraphics[width=0.49\linewidth]{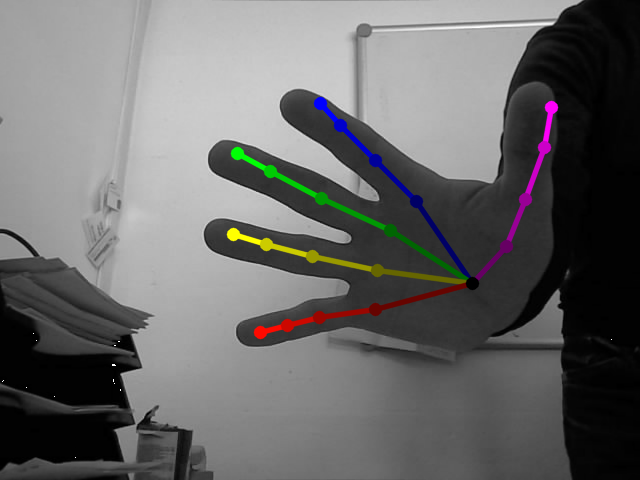}}
  \caption{Sample frames  of our two camera setup. We  capture hand
    articulations from a depth and an RGB camera and apply our  method on the depth
    camera to obtain 3D  annotations. These annotations are shown in (a).  To evaluate the accuracy, we project
    the estimated 3D annotations into the RGB camera,  which is  shown in
    (b). The full sequence is provided as supplemental material. (Best viewed in color)}
  \label{fig:two_cams}
\end{figure}

To also  evaluate real data,  we tested  our method on a calibrated camera setup
consisting of  a depth  camera  and an  RGB camera capturing  the hand  from two
different perspectives.   We create  3D annotations using  the depth  camera and
project them into the RGB camera.  We can then visually check the projections of
the annotations.  Fig.~\ref{fig:two_cams} shows one example: The joint locations
project nearby the real joint locations, which indicates that not only the image
coordinates are correct, but also the depth.  We refer to
the supplemental material for the full sequence.

\subsection{Application to the MSRA Dataset}

We applied our  approach to the MSRA dataset~\cite{Sun2015},  which is currently
the largest  dataset for hand  pose estimation  from single depth  images.  The authors
used  a state-of-the-art  3D  model-based method~\cite{Qian2014}  to obtain  the
annotations.  As discussed  earlier, these annotations are not  perfect. We used
our method to select  $10\%$ of the frames (849 out of 8499 for the first subject) as reference frames
and manually provided the 2D locations, visibility, and z-order of the joints for these
reference frames. It  took on average 45\,s per frame  for a non-trained annotator
to provide this information.
We further show a qualitative comparison, and that the higher annotation accuracy leads to better pose estimates, when training a state-of-the-art 3D hand pose estimator~\cite{Oberweger2015}.

\begin{figure*}
  \subfloat[Frame 970]{\includegraphics[width=0.12\linewidth]{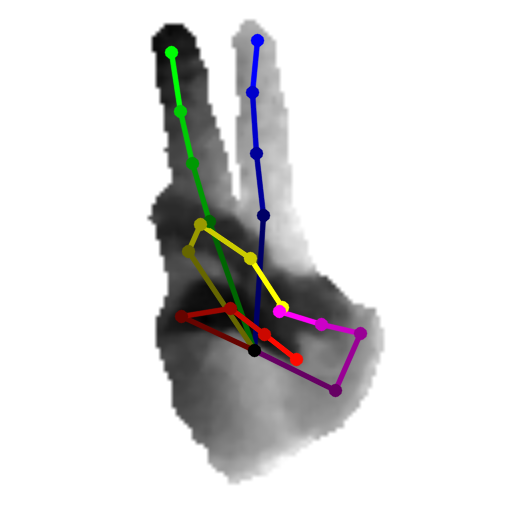}\includegraphics[width=0.12\linewidth]{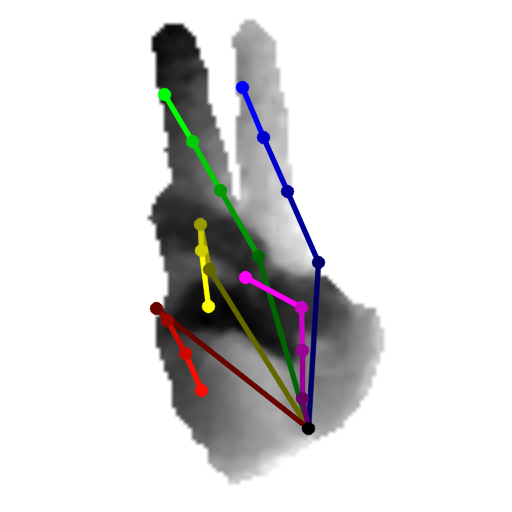}}
  \subfloat[Frame 1635]{\includegraphics[width=0.12\linewidth]{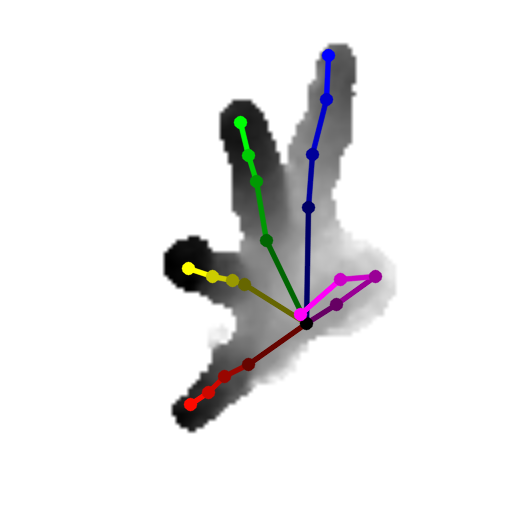}\includegraphics[width=0.12\linewidth]{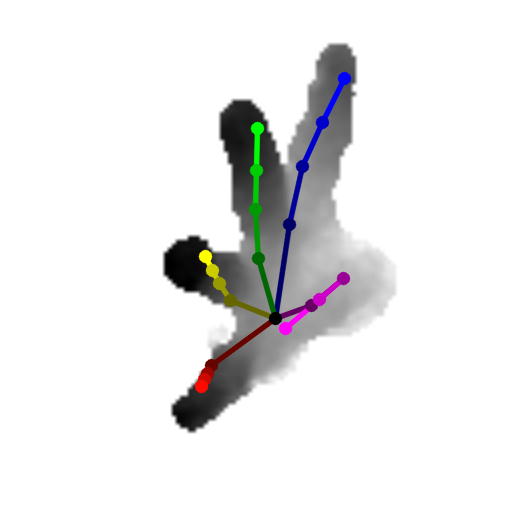}}
  \subfloat[Frame 3626]{\includegraphics[width=0.12\linewidth]{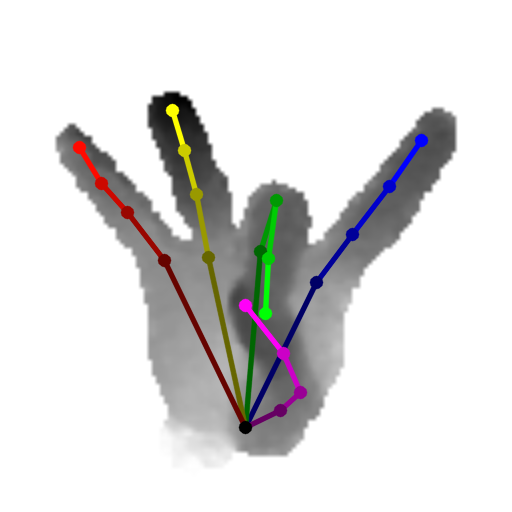}\includegraphics[width=0.12\linewidth]{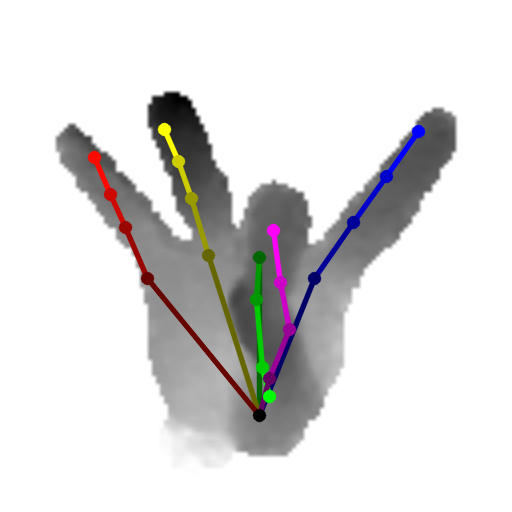}}
  \subfloat[Frame 8495]{\includegraphics[width=0.12\linewidth]{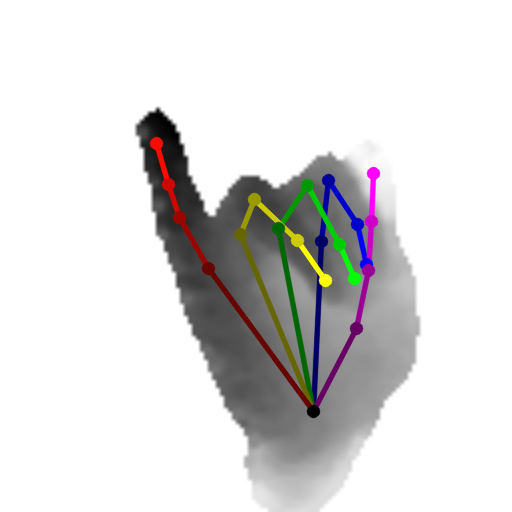}\includegraphics[width=0.12\linewidth]{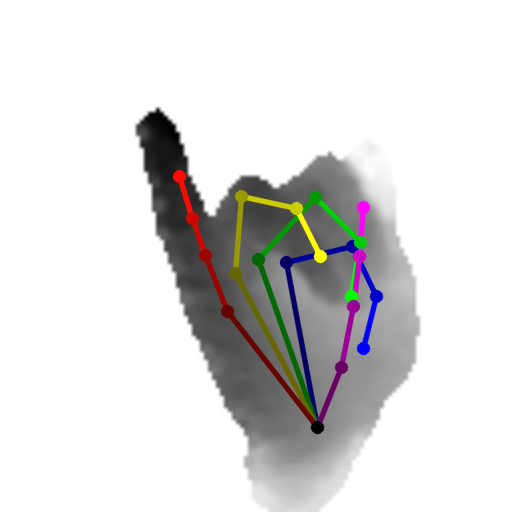}}
  \caption{Qualitative comparison  of the annotations obtained  using our method
    (left image) and the annotations of \cite{Sun2015} (right image) on the MSRA
    dataset.  We selected several frames  with large differences between the two
    annotations.  Note  that the shown  sample frames are  \emph{not} manually
    annotated reference frames. (Best viewed in color)}
  \label{fig:qualitative_eval}
\end{figure*}

\paragraph{Qualitative Comparison}
As ``real'' ground truth is not available, a direct evaluation is not
possible.  Fig.~\ref{fig:qualitative_eval} compares different
frames  for  which  the  distances between the annotations  are large.   Our  annotations  appear
systematically  better. This  strongly suggests  that our  annotations are  more
accurate over  the sequence. We provide  a video sequence that  compares the two
annotations as a supplemental material.

\paragraph{Higher Annotation Accuracy Leads to Better Pose Estimators}
We further show that better annotations improve the accuracy of state-of-the-art
3D hand  pose estimation methods.   We train the  method of~\cite{Oberweger2015}
with the  original annotations and compare  it with the estimator  trained using
the annotations we obtained with our method. For the evaluation, we perform 10-fold cross validation, because no explicit test set is specified~\cite{Sun2015}. The results of this experiment are
shown in Fig.~\ref{fig:train_iter}. The estimator trained with our annotations converges  faster, but to similar average joint  errors in the end.  This  indicates that training is easier when the  annotations are better. Otherwise, the estimator may focus on difficult, possibly wrongly annotated samples. The results clearly show that accurate training data is necessary to perform accurate inference. The estimator achieves test set errors of $5.58\pm 0.56$\,mm using our annotations for training and testing, and $6.41\pm 2.05$\,mm using the provided annotations. When we train the pose estimator on the provided annotations but evaluate it on our own annotations, the error is $13.23\pm 6.98$\,mm, which indicates discrepancy among the annotations. However, the visual comparison --- which is the best that can be done on real data --- shows that our annotations are more accurate. 

\bgroup
\setlength{\tabcolsep}{0pt}
\begin{figure}
  \begin{center}
\begin{tabular}{@{}lccc@{}}
\multicolumn{1}{R{90}{1em}}{Our anno} &
\includegraphics[width=0.33\linewidth]{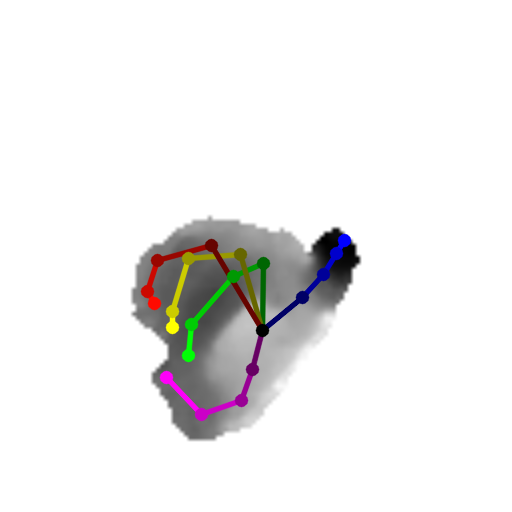} & 
\includegraphics[width=0.28\linewidth]{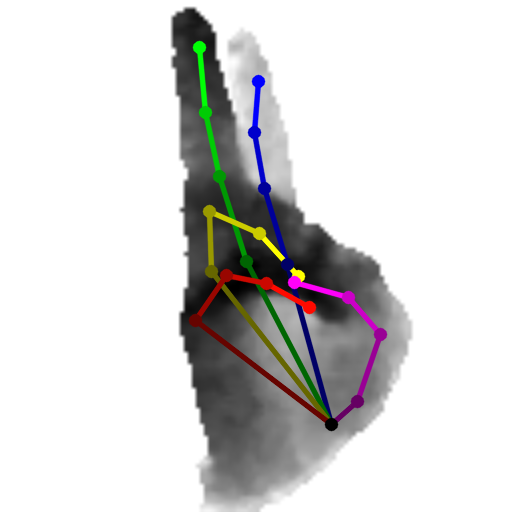} &
\includegraphics[width=0.30\linewidth]{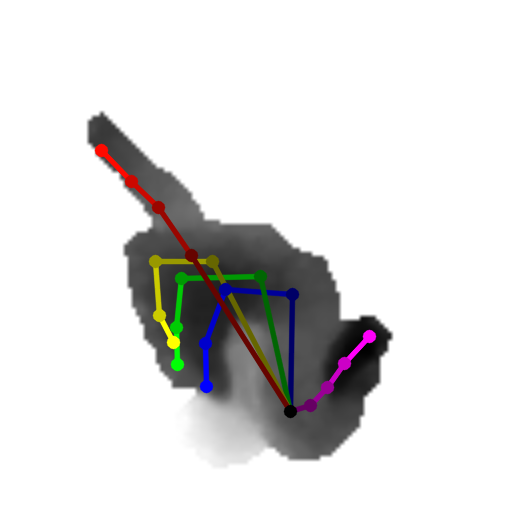} \\

\multicolumn{1}{R{90}{1em}}{Anno of~\cite{Sun2015}} &
\includegraphics[width=0.33\linewidth]{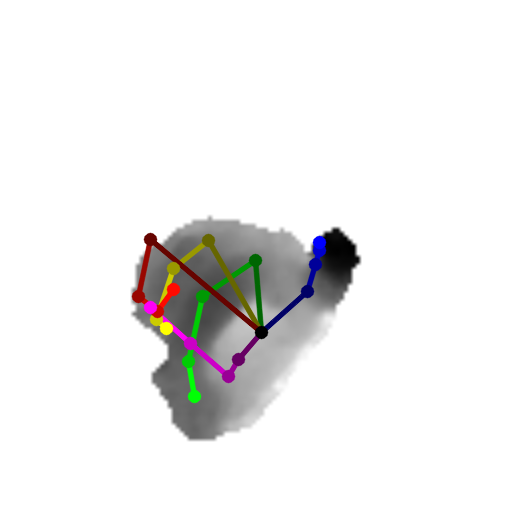} &
\includegraphics[width=0.28\linewidth]{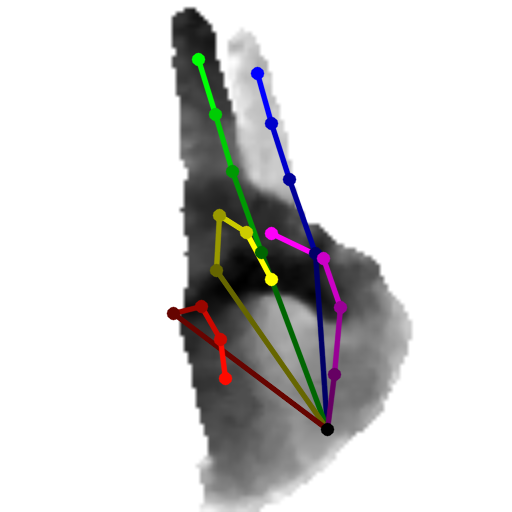} &
\includegraphics[width=0.30\linewidth]{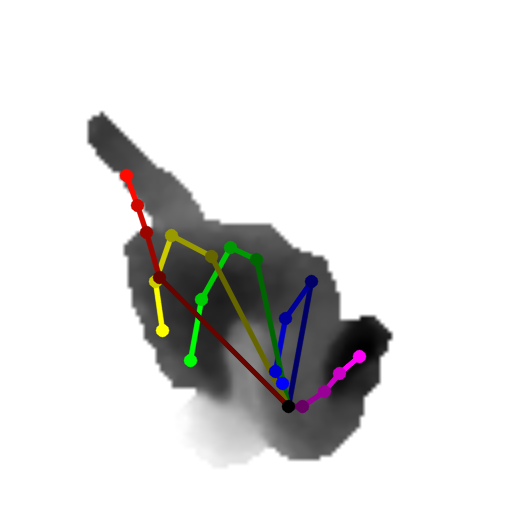} \\

& (a) & (b) & (c)\\
\end{tabular}
  \end{center}
  \caption{Training  and  testing  a  3D  hand  pose  estimator  with  different
    annotations for the MSRA dataset.  We  train a state-of-the-art 3D hand pose
    estimator~\cite{Oberweger2015} with  the provided  annotations and  with our
    revised annotations.   We then compare  the predicted annotations on  a test
    set. The  predictions with our annotations  are on the left  side. It  shows
    that  the  estimator  learns  the  incorrect  annotations,  which  leads  to
    inaccurate  locations  for  the  test  samples.   Using  our  more  accurate
    annotations leads to  more accurate results.  In (a) note  the annotation of
    the thumb, in (b) the annotations of  the pinky and ring fingers, and in (c)
    the articulation of the index finger. (Best viewed on screen)}
  \label{fig:train_iter}
\end{figure}
\egroup

\subsection{New Egocentric Dataset}
Egocentric 3D hand pose estimation is an appealing feature for different Augmented Reality or human computer interaction applications.
Creating  datasets for this task  is  very  difficult~\cite{Rogez2014}. Egocentric views show severe self-occlusions as fingers are often occluded by the hand and egocentric cameras have a limited field-of-view. Both facts result in a less reliable tracking. Even with manual initialization, fingers are frequently occluded and the hand can move outside the camera view frustum.

We provide  a new dataset consisting of more than 2000 frames  of several  egocentric sequences, each  starting and
ending with a  neutral hand pose and showing a user  performing a single or
various   hand  articulations   per   sequence.  We annotated the dataset using our  method. In contrast, the
3D model-based implementation of~\cite{Tompson2014} often failed  by converging  to different
local  minima, and  thus  resulted in  time consuming  fiddling  with the  model
parameters.

We  establish  a baseline  on  this dataset,  by
running   two   state-of-the-art   methods:    (1)   the   method   of
Oberweger~\etal~\cite{Oberweger2015}, which was shown to be among the
best      methods      for       third      person      hand      pose
estimation~\cite{Supancic2015},    and     (2)    the     method    of
Supan\v{c}i\v{c}~\etal~\cite{Supancic2015}, which  was initially  proposed for
hand pose estimation,  but especially for hand  object interaction in
egocentric views. We  perform   5-fold
cross-validation and  report the  average and standard  deviation over
the  different folds. We
report  the  results  in Table~\ref{tab:ego_results}.  The  method  of
Supan\v{c}i\v{c} has larger errors, mostly due to flipping ambiguities. For the oracle, we assume the 3D poses known, and return the pose with the smallest Euclidean distance from the training set.

\begin{table}
\begin{center}
  \begin{tabular}{lcc}
    \toprule
    Method & Avg. / median error\\
    \midrule
    Oberweger~\etal~\cite{Oberweger2015} & $24.58\pm 16.08$ / 19.53\,mm\\
    Supan\v{c}i\v{c}~\etal~\cite{Supancic2015} & $33.09\pm 21.66$ / 26.20\,mm\\
    Oracle & $20.20\pm 10.92$ / 19.47\,mm \\
    \bottomrule 
  \end{tabular} 
  \caption{Average accuracy on the egocentric hand dataset with 5-fold cross validations. We apply two state-of-the-art methods to the dataset and report the Euclidean 3D joint errors. For the orcle, we calculate the distance to the nearest sample in the training set.}\label{tab:ego_results}
\end{center}
\end{table}


\section{Conclusion}

Given the recent developments in Deep Learning, the creation of training data 
may now be the main  bottleneck in practical  applications of Machine Learning for hand pose estimation.
Our  method  brings a  much  needed  solution to  the  creation  of accurate  3D
annotations  of hand  poses. It avoids the  need for motion
capture systems, which are cumbersome and cannot always be used, and does not require complex camera setups. Moreover,  it could  also be  applied to  any other
articulated structures, such  as human bodies. 

\paragraph*{Acknowledgements:} 
This work was funded by the Christian Doppler Laboratory for Handheld Augmented Reality and the Graz University of Technology FutureLabs fund.

{\small
  \bibliographystyle{ieee}
  \bibliography{hand_dataset}
}

\end{document}